# RAISE – Radiology AI Safety, an End-to-end lifecycle approach

M. Jorge Cardoso *
School of Biomedical Engineering and Imaging Sciences, King's College London, London, UK

Julia Moosbauer *
deepc GmbH, Munich, Germany;
Department of Statistics, Ludwig Maximilian University Munich, Munich, Germany

Tessa S. Cook
Department of Radiology, Perelman School of Medicine at the University of Pennsylvania, Philadelphia, PA, USA

B. Selnur Erdal,
Center for Augmented Intelligence in Imaging, Department of Radiology, Mayo Clinic, Jacksonville, Florida, USA

Brad Genereaux
NVIDIA, Santa Clara, CA, USA

Vikash Gupta
Center for Augmented Intelligence in Imaging, Department of Radiology, Mayo Clinic, Jacksonville, Florida, USA

Bennett A. Landman
Vanderbilt University, Nashville, Tennessee, USA

Tiarna Lee
School of Biomedical Engineering and Imaging Sciences, King's College London, London, UK

Parashkev Nachev
UCL Queen Square Institute of Neurology, University College London, London, UK

Elanchezhian Somasundaram
Department of Radiology, Cincinnati Children's Hospital Medical Center, Cincinnati, OH, USA;
Department of Pediatrics, University of Cincinnati College of Medicine, Cincinnati, OH, USA

Ronald M. Summers
Imaging Biomarkers and Computer-Aided Diagnosis Laboratory, Radiology and Imaging Sciences, NIH Clinical Center, Bethesda, Maryland, USA

Khaled Younis
Phillips Research North America, Cambridge, MD, USA

Sebastien Ourselin *
School of Biomedical Engineering and Imaging Sciences, King's College London, London, UK

Franz MJ Pfister *
deepc GmbH, Munich, German

* Equal contribution

# Abstract

The integration of AI into radiology introduces opportunities for improved clinical care provision and efficiency but it demands a meticulous approach to mitigate potential risks as with any other new technology. Beginning with rigorous pre-deployment evaluation and validation, the focus should be on ensuring models meet the highest standards of safety, effectiveness and efficacy for their intended applications. Input and output guardrails implemented during production usage act as an additional layer of protection, identifying and addressing individual failures as they occur. Continuous post-deployment monitoring allows for tracking population-level performance (data drift), fairness, and value delivery over time. Scheduling reviews of post-deployment model performance and educating radiologists about new algorithmic-driven findings is critical for AI to be effective in clinical practice. Recognizing that no single AI solution can provide absolute assurance even when limited to its intended use, the synergistic application of quality assurance at multiple levels - regulatory, clinical, technical, and ethical - is emphasized. Collaborative efforts between stakeholders spanning healthcare systems, industry, academia, and government are imperative to address the multifaceted challenges involved. Trust in AI is an earned privilege, contingent on a broad set of goals, among them transparently demonstrating that the AI adheres to the same rigorous safety, effectiveness and efficacy standards as other established medical technologies. By doing so, developers can instil confidence among providers and patients alike, enabling the responsible scaling of AI and the realization of its potential benefits. The roadmap presented herein aims to expedite the achievement of deployable, reliable, and safe AI in radiology.

# 1. Introduction

The safe, effective, and ethical deployment of AI systems in radiology requires comprehensive efforts spanning the entire AI product lifecycle. Both a recent US President's Executive Order on the Safe, Secure, and Trustworthy Development and Use of Artificial Intelligence [1] and the EU AI Act [2] have further highlighted the importance of this topic in high impact areas such as healthcare. Healthcare AI already has strong regulatory guidance, with entities such as the Food and Drug Administration (FDA) and European Medicines Agency (EMA) having stringent guidelines governing clinical AI given its direct impact on patient health and safety [3]. Premarket testing and validation requirements, for example, aim to minimize risks from inaccurate or biased models, while post-market surveillance continuously monitors performance during production deployment. Meeting regulatory standards is thus already a legal obligation for healthcare AI.

*Primum non nocere* is one of principal precepts of bioethics and guides medical practitioners to first do no harm. With the increasing use of AI for patient care, it is imperative that models are bound to the same standards by avoiding preventable errors that could worsen outcomes. The moral mandate to equitably serve all patients, regardless of background, motivates minimising algorithmic errors while contributing towards building appropriate user trust; thus, both deontological and utilitarian ethics necessitate healthcare AI safety, effectiveness and efficacy.

AI models, which form part of software as a medical device (SaMD) solution, should be developed with inbuilt explicit (e.g. bias and fairness) or implicit (e.g. explainability) safety mechanisms. By making downstream users privy to these factors, AI developers aim to mitigate concerns and minimize risks inherently associated with using these models in a live scenario. Many SaMD solutions, however, do not provide such mechanisms; they often rely on the confidence in their large development datasets to model the target patient population distribution relatively well, and software environment assumptions. This is due to the complexity and cost of implementing systematic pre-deployment quality control and quality assurance procedures, live monitoring systems, and scheduled post-deployment surveillance. With nearly 700 AI-based SaMD solutions [4] on the market, replicating this infrastructure on a per-solutions basis would be very complex due to cost and scalability. A practical solution to this scalability problem is to implement these quality, safety, and monitoring mechanisms at the AI orchestration platform level [5], ensuring trustworthy and safe AI is procured, commissioned, delivered, and integrated uniformly across all hospital AI solutions [6]. Such a platform approach would also centralise the assessment of mid- to long-term outcomes, enabling a better understanding of solution value.

Trustworthy and safe AI needs a holistic life-cycle approach, tackling pre-, peri- and post-deployment stages. In the pre-deployment stage, regulatory due diligence ensures models meet guidelines for analytic and clinical validity set by regulatory bodies such as the FDA, MHRA or EMA, and institutional AI governance councils. Extensive quality assurance testing of software allows for the identification of deployment time bugs and/or computational performance issues, while independent validation studies on diverse data can help identify potential model accuracy issues and generalisability limitations prior to deployment. During live deployment, input data quality needs to be closely monitored to ensure it matches the quality and type of data used during pre-deployment testing, while output guardrails such as uncertainty estimates, and human-in-the-loop reviews act as a redundant safety net allowing models to defer uncertain decisions to clinicians. Lastly, in the post-deployment phase, continued surveillance ensures models maintain their safety and effectiveness, that algorithm performance drift caused by changes in data or population characteristics are measured appropriately and models re-calibrated, and fairness and bias assessments proactively uncover inequities in model behaviour between patient subgroups.

This end-to-end approach to trustworthy AI deployment, summarized in Figure 1 below, emphasizes safeguards across the AI lifecycle, including regulatory, clinical, technical, and ethical considerations. When rigorously implemented, it fosters provider confidence in applying AI technologies to improve patient outcomes and quality of care. Institutional support and collaboration between diverse experts are crucial for navigating the sociotechnical complexities of AI in radiology.

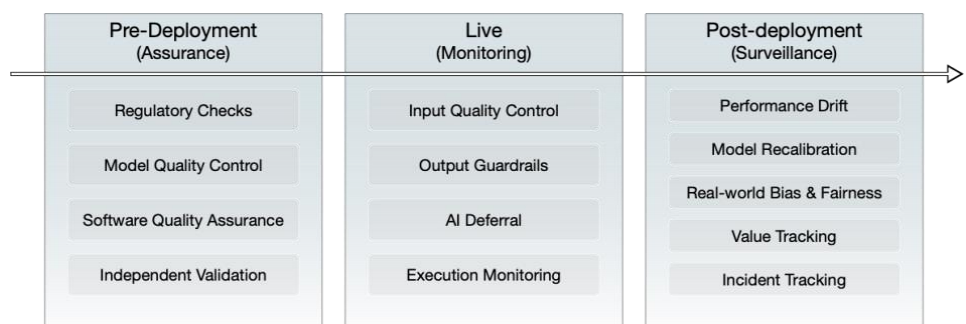

*Figure 1 – Diagram representing the proposed radiology AI safety lifecycle approach*

# 2. Pre-Deployment phase

This section will discuss pre-deployment AI safety, focusing on the need for regulatory due diligence, followed by a discussion around model quality control and software quality assurance, and finally highlighting the need for formal independent validation.

## 2.1. Regulatory Due Diligence

There are several key considerations when deploying radiology AI models in clinical settings to ensure patient safety and effective care, from statutory regulation to professional standards and general desirability. One of these is regulatory due diligence, i.e. ensuring AI models meet relevant regulations and standards. For example, the FDA or the EU's MDR, has specific guidelines for the approval of medical devices, including AI-based systems. These can be informally augmented by professional societies (e.g. American College of Radiology, Radiological Society of North America, European College of Radiology), promoting best practice recommendations for AI usage and reporting [7], [8]. Any entity implementing these SaMD solutions, from healthcare systems and hospitals to AI marketplaces, should conduct a thorough due diligence process to ensure all implemented models are safe, effective, and regulatory compliant.

The regulatory approval process for AI-based medical devices differs in complexity and rigor between countries, ranging from mostly procedural to mostly scientifically rigorous requirements, but all aim to validate the device development methodology, and algorithmic safety and efficacy through some form of premarket validation. Manufacturers must submit detailed information to regulators on a plethora of aspects such as intended use, algorithm design, performance testing, risk analysis, quality management systems, and post-market surveillance. Regulators reviewing submissions may request additional information, such as prospective clinical studies, to substantiate the performance of the AI model; approval is often based on reasonable assurances of safety and effectiveness for the intended use. Many research groups, with the support of professional societies such as ACR and RSNA, have enhanced formal regulatory requirements by publishing guidance documents such as STARD-AI [9], FUTURE-AI [10], MI-CLAIM [11] or MINIMAR [12], outlining principles and best practices for healthcare AI. These cover topics like appropriate clinical validation and their evidence hierarchy, transparent reporting of capabilities and limitations, and trial protocol design considerations. While informal, these guidelines represent a consensus expert view and can serve as complements to formal regulations.

AI deployment platforms also have an obligation to their users to ensure models listed on their platforms meet all regulatory requirements. Testing performed by platform providers should involve several crucial steps and should be geographically informed, as every market will have its own set of requirements. First, it requires confirming that the system holds the necessary certifications and clearances to operate legally within specific regions or countries, as regulatory requirements may differ across locations. Secondly, it necessitates compliance with registration obligations, by registering the AI system with relevant authorities or agencies. Lastly, verifying the availability of comprehensive user-facing materials in relevant languages is required for effective and safe utilization of the system by healthcare professionals. These verification steps collectively contribute towards a thorough due diligence process before deploying a radiology AI system.

## 2.2. Independent model validation

External and independent model validation, either as an independent retrospective data analysis for self-contained models, or preferably, a standalone prospective randomised controlled trial studying end-to-end solution value, can provide reassurances on model performance, calibration, value, and risk. By combining the findings of multiple smaller independent studies into a single meta-study, one can often gather more comprehensive evidence than the original validation set used for regulatory approval.

While regulatory agencies require that AI algorithms used in medical devices undergo rigorous testing and validation prior to regulatory approval for clinical use, a broadly comprehensive validation is hard to attain. This is especially important given that the clinical validation of AI models required for regulatory clearance is, in retrospect, often limited in scale and diversity. Many models are tested on curated limited datasets that may not fully capture real-world variability across different hospitals, scanners, and populations, motivating the need for supplemental external validation studies that can be designed to focus on diverse challenging cases missed during initial validation. The degree and hierarchy of evidence in AI validation studies is thus of great importance.

While independent validation studies can be conducted by AI-deploying hospitals, research institutions, AI marketplace companies, or medical societies, meta-analyses of multiple small studies can provide additional performance insights. The aggregated results of a meta-study can identify a model's particular areas of struggle, providing insights towards generalizability and enabling the users to better understand the appropriate usage context and limitations of an AI model. It is recommended that AI model and platform vendors streamline the process of independent validation by potential client institutions, both via purpose-specific platforms and by enabling broad data access, e.g. via federated learning [13]. Standardized technical support for quick deployment and monitoring during evaluation trials with transparent data use agreements can be beneficial for all parties.

Insights from these independent validation efforts should even be shared with regulatory bodies to aid in continued oversight and to build the body of evidence needed for safe AI usage. Model manufacturers also benefit from the feedback to improve their algorithms or deployment recommendations. Most importantly, such supplemental validation, primarily when used in a continuous and ongoing manner, helps maintain trust and confidence in the AI model among end-users.

## 2.3. Model Quality Control

Another important aspect of AI deployment is to ensure models perform well at each hospital they are deployed during the commissioning process. This includes validating and documenting the accuracy and reliability of a model using hospital and population-specific real-world datasets.

Often AI models are developed and tested using limited datasets that may not fully represent real-world diversity [14]. Hence, locally validating models before deployment is crucial. Ideally, one should curate a representative dataset from the hospital covering different patient demographics, scanner models, pathologies, etc, with the AI model then being evaluated on this dataset to gauge real-world performance. This is particularly important as disease prevalence and phenotype can change significantly worldwide, *e.g.* lung nodule subtype prevalence differs significantly between countries and regions. [15]

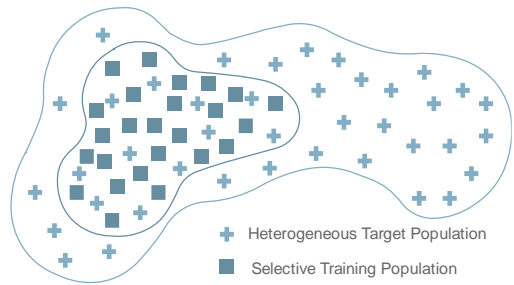

*Figure 2 - Diagram depicting differences in population spread between classical high-quality selected training that matches inclusion criteria, and highly heterogeneous unselected target populations with complex clinical presentations.*

Quantitative metrics like accuracy, sensitivity, specificity, AUC, as well as a qualitative clinical review of model outputs, are important at the pre-deployment quality control (QC) stage. A site-specific QC step also allows the evaluation of different operating thresholds, which can optimise trade-offs between metrics (e.g. sensitivity/specificity) for a target population.

It is important to note that this quality control check can be either algorithmic or based on human experts, and should target model inputs (e.g. does the input data look like what is expected), model outputs (i.e. do the model outputs make sense given an input), and population level statistics (i.e. is the model well calibrated at the population level). While human-based quality control for all AI models and sites would be ideal, it is often non-scalable due to limited clinical time availability and cost. New algorithmic solutions for input quality control [16], input domain shift [17], output quality [18] and model calibration [19] will help standardize, automate, and make this QC step more scalable.

**2.4. Software quality assurance**

The software technology stack used to develop radiology AI models must be of high quality and meet certain standards for safety and efficacy, ideally building upon trusted open-source toolkits such as MONAI [20]. When AI medical device manufacturers package these AI models into a complete SaMD, they need to ensure, to the best of their ability, that their software is free from errors, bugs, and security vulnerabilities, and that it can handle large amounts of data. While most quality assurance (QA) processes are implemented by the manufacturer itself, bugs and errors in AI models are often a function of the input data, and thus models need to be continuously QA tested at every site they are deployed to ensure that AI models are reliable and effective.

Like any other software, new errors and defects can appear during installation and deployment. Common issues, such as unexpected algorithmic failures (e.g. high-resolution image causes failure due to lack of compute memory), incorrect inputs (e.g. unsupported data geometry), and slow performance (e.g. particular data taking too much time to process) require rigorous software QA testing to identify these before deployment. While manufacturers conduct QA during development, additional testing (in real-world settings) should be done after integrating the software with the hospital IT systems. Differences in infrastructure, data formats, and usage patterns can cause new issues to surface. QA testing the deployment platform with real hospital data at scale is advisable before live availability. It is also important to load, and stress test all models to evaluate their performance and scalability under heavy usage, and to have a clearly defined process for resolution based on issue severity and associated service level agreements.

# 3. Production deployment monitoring phase

This section will discuss different types of AI safety during production deployment, ensuring input and output data is appropriately controlled for, that algorithms can *know what they don't know*, and that algorithmic execution is appropriately monitored.

### 3.1. Input Quality Control

Input data quality is of paramount importance when ensuring AI models are safe and accurate. While data quality checks are often done by AI manufacturers themselves as part of their pre-deployment QA , most models depend on some system-level control to ensure data quality. For example, in a radiology practice, the imaging examination should be of the right modality, have the correct organ in the field of view and be of sufficient diagnostic quality for further downstream analysis so that it can be correctly processed by an AI model and generate the correct output. Real-time quality control mechanisms should be in place to ensure that input data distribution, or the task concept itself, does not drift from the one used during pre-deployment AI quality control, as model performance would not otherwise be guaranteed.

AI models are very sensitive to the quality and format of input data they receive. Issues with input data are a common cause of unreliable model performance after deployment. More specifically, live input quality can be monitored through:

- **Data type checking** - Confirm inputs match expected modalities (X-ray, CT, MRI etc.)
- **Metadata validation** - Data should have DICOM tags indicating the imaged body part. Inputs should match the model's intended use and anatomy.
- **Acquisition parameter checks** - Confirm protocol parameters like scanner acquisition characteristics, image resolution, and use of contrast are within specified bounds.
- **Quality assessment** - Measure noise levels, motion artifacts, foreign bodies, contrast levels etc. and check against prescribed threshold ranges from the AI vendor.

Both programmatic checks and human review are useful here: input quality issues can trigger alerts for technicians to rectify acquisition, and population shifts can require a new quality control step. Unsurprisingly, the same class of algorithms used to automate pre-deployment QC (sec 2.3) such as input quality control [Sadri et al. 202] and input domain shift [Graham et al. 2022 PMLR] can be repurposed for live deployment and continuous monitoring. Summarily, high-quality and trusted inputs are the bedrock for reliable AI model performance.

### 3.2. Output Guardrails

When an AI model is presented with data similar to the one the model has seen during training, performance should be comparable to the one observed during model development, as the model is "in-distribution". However, in order to further minimize AI risks at the individual patient level, some AI manufacturers implement additional output guardrail mechanisms in the form of model confidence or uncertainty estimation [Mehrtash et al. 2020 IEEE TMI], explicitly explainable models that can then be logically controlled for, or separate redundant quality control systems [Valindria et al. 2017 TMI]. However, as these output guardrail mechanisms are only seldom in place, a platform-level redundancy mechanism may be necessary to ensure safety across all deployed AI models. It is also important that these failures are appropriately recorded and communicated downstream so that patients can be safely cared for. While input quality control aims to ensure models receive appropriate data, output guardrails monitor the models' actual performance on those inputs, with approaches depicted in Fig 4 below, such as:

- **Uncertainty estimation** – Models predict an uncertainty score along with each output signifying its confidence. Outputs with high uncertainty can be flagged for review.
- **Consistency** - Passing the same input to multiple independent models (for the same task) and checking if their predictions agree can help detect errors.
- **Logical constraints** – Outputs are checked against logical rules and constraints, e.g. for an estimated volume being within an expected range. Violations are flagged.
- **Distribution monitoring** – The distribution of model outputs is tracked for drift over time. Sudden distribution shifts could indicate a problem.
- **Human-in-the-loop** – A certain percentage of outputs could be validated by human expert reviewers. Disagreements indicate poor performance.

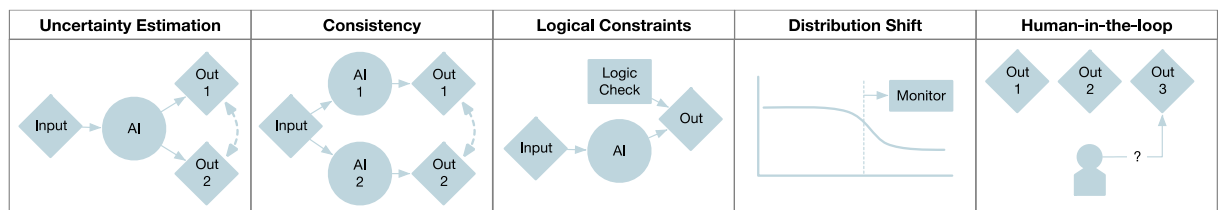

*Figure 4 - Depiction of different types of system level safety mechanisms*

The platform should define operating thresholds for these checks based on model specifics and risk tolerance. Detected issues should trigger alerts to the underlying AI medical device manufacturer, along with deferring the results from getting to clinical users. This allows for proactive debugging and re-training of models if needed.

### 3.3. Deferral of Algorithmic Decisions

With the introduction of platform-level input and output guardrail mechanisms, it is important to ensure these acts as a fail-safe. If an algorithmic output is deemed not to be of sufficient quality according to a locally calibrated threshold, it should be deferred to the most appropriate downstream expert clinician while providing a transparent explanation for the deferral reason. It is important to note that while deferral mechanisms do not fully mitigate but minimise risk in a medical device. Such a deferral mechanism should also be integrated with the patient's electronic health record or the radiological information system to ensure that all patients are appropriately cared for regardless of the reason that triggered the deferral.

When model outputs fail guardrail checks, they cannot be shown to users and must be deferred to human experts for review. However, deferring directly can lead to patient care delays, gaps in the workflow, or additional work burden on clinical staff. The deferral mechanism should therefore seamlessly integrate with existing clinical reporting systems, adhere to interoperability standards and profiles, and automatically assign deferred cases to and notify radiologists based on configurable rules accounting for sub-specialization, availability etc. It should also track deferred case workload at the platform and individual user levels for analytics and provide auditable logging of all deferrals with clear reasons linked to the patient record. With tight integration into existing systems and clinical workflows, deferred cases due to guardrail failures can be handled gracefully avoiding gaps in patient care.

### 3.4. Model Execution Monitoring

Lastly, as AI models are ultimately software, it is important that we track key performance indicators related to model runtime, failure and deferral rates, accuracy, precision, and recall, so these can be audited and managed. If an unexpected failure occurs, or the execution time of the models is beyond the expected service level agreement, a system administrator needs to be notified via an incident management system. It is also important that all model executions are appropriately logged in a harmonized and well-documented manner for post-hoc analysis and quality assurance purposes. In addition to monitoring model inputs and outputs, tracking technical metrics around the model execution itself is also important:

- **Model versioning** - Link executions to specific model versions to track provenance.
- **Runtimes** - Monitor latency, throughput and model inference times of the AI application, and flag violations of service agreements.
- **Hardware/resource utilization** - Track GPU/RAM usage spikes indicating performance issues.
- **Errors/failures** - Log any unexpected errors like GPU out-of-memory failures.
- **Audit logs** - Detailed execution logging for traceability and recreation of issues.
- **Deferral rates** - Measure overall deferral rates as well as reason-wise breakdown.
- **Accuracy metrics** - Log model outputs and eventual clinician diagnoses and dissent to track accuracy over time.

These metrics provide insights into model performance in production and can alert to regressions, and debugging data enables root cause analysis of problems by AI teams as long-term metric trends can identify needs for re-training or upgrade. Detailed logs are also crucial for regulatory compliance and audits. Execution monitoring, together with the input and output guardrails, provides full and comprehensive visibility into AI systems.

# 4. Post-market surveillance phase

### 4.1. Performance Drift Monitoring
Algorithmic guardrails and monitoring during live AI deployment allow the identification of subject-level failures and problems, but many effects are only visible at the population level. For example, algorithmic performance might decay over time due to changes in how the data are acquired, updating of imaging scanners and sequences, or phenotype/population drift effects. It is thus important to track and audit model performance; this can be achieved algorithmically, via manual clinical review, or by comparing with downstream clinical diagnosis and outcomes as recorded in the electronic health record. If performance degradation is observed, a full clinical review of the problem, and subsequently redoing the quality control study, will be necessary to avoid further impacting patient care.

AI model performance can slowly drift from validated levels post-deployment due to a variety of reasons, from data drift, i.e. gradual changes in input data characteristics like resolution, contrast, or new scanners/protocols; to phenotype drift, i.e. changes in disease morphology and presentation over time; and concept drift, i.e. evolution in best practices and diagnostic standards. It is also important to note that deploying AI might create degenerate feedback loops between AI recommendations and outcomes [21], further drifting performance.

Continuously monitoring metrics like accuracy, sensitivity, and specificity versus time can detect such gradual performance degradation. Comparing model outputs against clinician diagnoses over a period of time provides another signal of drift.

Upon suspected drift, a retrospective analysis of model errors should be undertaken. Additional blinded validation by clinical experts on a recent dataset would provide a gold standard quantification of the extent of degradation. Based on the findings, the model may need retraining on suitable additional data, and adjustment of operating thresholds, or model retirement. Being proactive about monitoring and addressing drift is crucial to prevent a gradual decline in model utility over time.

### 4.2. Model Recalibration

Similarly, to performance drift, algorithms often make predictions according to either development-stage or pre-deployment stage calibration of the operating point of a receiver-operating curve (ROC), targeting a specific sensitivity or specificity depending on the clinical use case. For example, screening and assessment stages of breast cancer diagnosis requires different sensitivity/specificity operating points. As data and models drift, it is often necessary to recalibrate the ROC curve operating point to ensure the clinically ideal behaviour of the model, which is a function of regional and/or physician preference. By tracking population-level metrics over time and conducting audits, degrading sensitivity or specificity can be identified. The current threshold and desired levels can be provided to AI teams to re-calibrate the models accordingly. Even though this process can be automated [cite ben], it is important to note that fully automated continuous recalibration is currently infeasible in certain geographies due to regulations. However, having processes to periodically adjust thresholds in consultation with medical leaders ensures optimal clinical utility. Overall, tracking algorithmic sensitivity and specificity over time and triggering a review process is necessary to ensure AI models behave appropriately.

### 4.3. Bias and Fairness

AI safety is often perceived as either a population-wide or a subject-specific task, but several harmful algorithmic effects can occur in subsets of the overall population. From a bias and fairness point of view, and if AI is to be trusted by all members of society, it is important to verify that model performance is equitable across different population subgroups, primarily between clusters of subjects that share protected characteristics. If sufficient evidence of sub-population underperformance is found (including both protected characteristics and phenotypical variance), a full clinical review should be conducted to assess the impact on those populations, followed by an appropriate algorithmic remediation and notification of both manufacturer and regulators of this adverse finding.

While overall metrics may show a model is safe, effective and efficacious, it can still disproportionately underperform in certain population segments. Unfairness can be caused by underrepresentation in training data, where a model may not have seen enough examples from that population to generalise well, proxies and correlations, where a model may learn to depend on covariates that correlate with sensitive attributes, and concept drift, where disease patterns and clinical standards for different populations may differ. Remediation strategies can be introduced during pre-processing (e.g. targeted data collection), in-processing (e.g. adversarial learning), and post-processing (calibration).

It is also important to note that unfair model performance, if found, should warrant detailed disclosure to providers, patients and carers from that community, as collaborating with leaders from the community on mitigation approaches will contribute to building trust. Efforts to improve model equity should also be undertaken, including gathering more representative data.

### 4.4. Incident Tracking

Regulators increasingly require manufacturers to establish an incident tracking system as part of a post market surveillance system to proactively collect and review experience gained from devices that are placed on the market, with the aim of identifying any need to immediately apply corrective or preventive actions. A key component is maintaining an incident tracking process to record issues on device performance or safety, including model provenance, data hashes, hardware configuration and integration tooling, that come to the manufacturer's attention after market release. Incident data helps identify software anomalies, or use errors not caught during pre-deployment testing. Rapidly detecting and addressing incidents is not only a legal obligation, but also vital for managing risks and ensuring the AI technology remains safe and functions as intended post deployment.

### 4.5. Value Tracking

Value-based healthcare [22] is a model where providers are reimbursed based on the value of care they deliver, rather than the volume of services provided. AI safety is thus interlinked with value-based healthcare [23], not only because a simple AI model decision can have long-term implications in the outcome and thus the value of the care delivered to patients, but also because a better understanding of the value created by a model would allow for a more informed view of when and to which degree a model should be used in a clinical setting. There also is a disparity between vendor and patient incentives, as, for example, an AI vendor might prefer a higher case throughput while a specific patient might not benefit from the increased cost. Overall, while a better understanding of the value of an AI model does not necessarily contribute towards AI safety, it does significantly contribute towards trust in the technology itself, primarily from the point of view of patients and their carers.

As value-based care focuses on optimizing patient outcomes while lowering costs, if AI is to be made safe but also trustworthy, it is important to capture the different sources of the AI value chain, such as enabling earlier and more accurate diagnoses leading to timely intervention, reducing unnecessary diagnostic testing in obvious negative cases, and automating tedious workflows improving clinician productivity. However, unsafe or poorly performing AI can negatively impact value through inaccurate diagnoses, prognosis, planning, among other use cases, resulting in improper treatment plans, loss of trust in AI leading to ignoring valid model outputs by clinicians, and liability costs and reputational damage from adverse events involving AI. Evaluating the impact of AI on metrics like clinical outcomes, patient satisfaction, clinical workflow integration, operational efficiency, and costs is crucial. Overall, a virtuous cycle of deploying AI safety initiatives to build provider trust is needed, thus enabling expanded use of AI, and leading to greater value delivery.

# 5. Conclusion

The deployment of AI systems in radiology holds great potential to improve clinical care, but also carries risks if not implemented thoughtfully. As outlined in this paper, a holistic approach to trustworthy AI that encompasses the entire product lifecycle is crucial.

Rigorous pre-deployment evaluation and validation ensure AI models are safe and effective for the intended use case. Input and output guardrails during live usage provide an additional safety net to identify individual failures. Continuous monitoring after deployment enables tracking of population-level performance, fairness, and value delivery over time.

No one solution alone can guarantee absolute safety. Rather, it is the synergistic application of quality assurance at multiple levels - regulatory, clinical, technical, and ethical - that fosters appropriate trust in AI. Collaboration between stakeholders across healthcare systems, industry, academics, and government is vital to address the interdisciplinary challenges involved.

Trust is earned, not given. By transparently demonstrating that AI is held to the same safety and efficacy standards as other medical technologies, developers can build confidence among providers and patients. This will enable the responsible scaling of AI and realization of its potential benefits. The roadmap presented here aims to accelerate this goal of deployable, reliable, and safe AI in radiology.

**Funding/Disclosures**

M. J. Cardoso and S. Ourselin are funded by the Wellcome Programme on high-dimensional neurology (WT213038/Z/18/Z) and Wellcome EPSRC CME (WT203148/Z/16/Z). T. Lee is funded by the Engineering & Physical Sciences Research Council Doctoral Training Partnership (EPSRC DTP) grant EP/T517963/1. R. M. Summers receives royalties from iCAD, Philips, ScanMed, PingAn, Translation Holdings and MGB. He has received research support from PingAn. This work is funded in part by the Intramural Research Program of the NIH Clinical Center. The opinions expressed herein are those of the authors and not necessarily those of the U.S. Department of Health and Human Services or the National Institutes of Health.